\documentclass[conference]{IEEEtran}
\usepackage{algorithm}
\usepackage{algorithmic}

\usepackage{graphicx}
\usepackage{textcomp}
\usepackage{xcolor}
\usepackage{cite}

\usepackage{amsmath}
\usepackage{amssymb}
\usepackage{amsthm}
\usepackage{MnSymbol}
\usepackage{textcmds}
\usepackage{stmaryrd}
\usepackage{tabularx,makecell}
\usepackage{hyperref}

\usepackage{tikz}

\newtheorem{definition}{Definition}

\newtheorem{theorem}{Theorem}
\newtheorem*{theorem-nonum}{Theorem}
\DeclareMathOperator*{\relu}{ReLU}
\DeclareMathOperator*{\argmax}{arg\,max}
\DeclareMathOperator*{\relumax}{relu\,max}

\usepackage{cleveref}

\newif\ifPreprintCopyright
\PreprintCopyrighttrue
\newcommand\copyrighttext{%
  \footnotesize \textcopyright 2021 IEEE. Personal use of this material is permitted.  Permission from IEEE must be obtained for all other uses, in any current or future media, including reprinting/republishing this material for advertising or promotional purposes, creating new collective works, for resale or redistribution to servers or lists, or reuse of any copyrighted component of this work in other works.}
\newcommand\copyrightnotice{%
\begin{tikzpicture}[remember picture,overlay]
\node[anchor=south,yshift=10pt] at (current page.south) {\fbox{\parbox{\dimexpr\textwidth-\fboxsep-\fboxrule\relax}{\copyrighttext}}};
\end{tikzpicture}%
}

\begin{document}
\title{Geometric Path Enumeration for Equivalence Verification of Neural Networks
%{on Structurally Different Neural Networks}
}

\author{\IEEEauthorblockN{Samuel Teuber\IEEEauthorrefmark{1}, Marko Kleine Büning\IEEEauthorrefmark{2}, Philipp Kern\IEEEauthorrefmark{3} and Carsten Sinz\IEEEauthorrefmark{4}}
\IEEEauthorblockA{Department of Theoretical Computer Science\\
Karlsruhe Institute of Technology (KIT), 
Germany\\
Email: \IEEEauthorrefmark{1}\url{samuel@samweb.org}, \IEEEauthorrefmark{2}\url{marko.kleinebuening@kit.edu},
\IEEEauthorrefmark{3}\url{philipp.kern@kit.edu}, 
\IEEEauthorrefmark{4}\url{carsten.sinz@kit.edu}}}

% \author{\IEEEauthorblockN{Samuel Teuber%\,\orcidlink{0000-0001-7945-9110
% \textit{Karlsruhe Institute of Technology}\\
% Karlsruhe, Germany \\
% \and
% \IEEEauthorblockN{Marko Kleine Büning}
% \textit{Karlsruhe Institute of Technology}\\
% Karlsruhe, Germany \\
% \and
% \IEEEauthorblockN{Philipp Kern}
% \textit{Karlsruhe Institute of Technology}\\
% Karlsruhe, Germany \\
% \and
% \IEEEauthorblockN{Carsten Sinz}
% \textit{Karlsruhe Institute of Technology}\\
% Karlsruhe, Germany \\
% \and
% \and

\maketitle
\ifPreprintCopyright
\copyrightnotice
\fi
\pagestyle{plain}

\begin{abstract}

% An increasing trend in applying neural networks to safety-critical domains necessitates their formal verification.
% One important property in this context is neural network equivalence,
% %One branch of neural network verification is concerned with equivalence verification.
% because networks are oftentimes compressed to fit computational and space restrictions of embedded systems.
% %, there is a need to prove such networks equivalent.
% %For the practical deployment of such networks empirical correctness as provided by those approaches is not enough.
% Existing techniques for the equivalence verification of neural networks are either based on exact MILP encodings, which lack scalability, or interval propagation, which is impractical for structurally different or retrained networks.  

% First, we provide a proof that the $\epsilon$-equivalence problem for neural networks is coNP-complete, then we present \textsc{NNEquiv} --
% a novel technique to verify the equivalence of neural networks which extends Tran et al.'s geometric path enumeration approach.
% We optimize our method through over-approximations as well as branching and refinement heuristics. 
% %Optimizations of utilized data-structures, LP-problems and branching decisions increase the scalability of our approach.
% An evaluation on ACAS Xu and modified MNIST benchmarks shows that \textsc{NNEquiv} can prove equivalence
% in cases where previous approaches like \textsc{MilpEquiv} and \textsc{ReluDiff} time out.

As neural networks (NNs) are increasingly introduced into safety-critical domains, there is a growing need to formally verify NNs before deployment.
In this work we focus on the formal verification problem of NN equivalence which aims to prove that two NNs (e.g. an original and a compressed version) show equivalent behavior.
Two approaches have been proposed for this problem:
Mixed integer linear programming and interval propagation.
While the first approach lacks scalability, the latter is only suitable for structurally similar NNs with small weight changes.

The contribution of our paper has four parts.
First, we show a theoretical result by proving that the epsilon-equivalence problem is coNP-complete.
Secondly, we extend Tran et al.'s single NN geometric path enumeration algorithm to a setting with multiple NNs.
In a third step, we implement the extended algorithm for equivalence verification and evaluate optimizations necessary for its practical use.
Finally, we perform a comparative evaluation showing use-cases where our approach outperforms the previous state of the art, both, for equivalence verification as well as for counter-example finding.

\end{abstract}


\section{Introduction}
With the success of deep neural networks (NNs) in recent years, there has been an increasing trend to introduce machine learning based approaches into many domains -- including safety-critical areas,
such as airborne collision avoidance~\cite{julian2016policy} and autonomous cars~\cite{bojarski2016end}.
%With the success of deep neural networks in recent years, there has been an increasing trend to introduce machine learning based approaches into many domains -- including domains which are classically considered as safety-critical.
%For instance, such domains include airborne collision avoidance ~\cite{julian2016policy} and autonomous car driving systems~\cite{bojarski2016end}.
Therefore, 
%As neural networks are introduced into such domains, 
there is a growing interest in verification of NNs.
This spurred research on new verification methods~\cite{katz2017reluplex,Singh2018,wang2018efficient,tran2019star,Buning2020,paulsen2020neurodiff,narodytska2018verifying}.
%\todo{In particular: VNN Competition \cite{VNN}}
The literature can broadly be classified into adversarial robustness verification (e.g. the work by Singh et al.~\cite{Singh2018}), functional property verification (e.g. the work by Katz et al.~\cite{katz2017reluplex} on the verification of ACAS Xu) and equivalence verification (e.g. Kleine Büning et al.~\cite{Buning2020} and Paulsen et al.~\cite{paulsen2020neurodiff}).
%We will refer to the mixed integer linear programming (MILP) based approach by \cite{Buning2020}, as \textsc{MilpEquiv}.
%While the first category is concerned with a problem specific to NNs, i.e. the question whether there are seemingly ``normal'' inputs which produce unexpected outputs, the second and third category are verification tasks known from ``classical'' work on software verification:
%They deal with the question whether a given NN fulfills certain functional properties (functional verification) or whether two NNs are equivalent.
%Functional verification is particularly relevant to ensure that a NN is capable of handling a certain safety critical task.
The main application of equivalence verification %, however,
is in the space of NN compression.
As NNs grow ever larger and computing becomes ever more ubiquitous,
resource restrictions require to compress large NNs into smaller models.
Cheng et al.~\cite{cheng2020survey} give an extensive survey of such compression techniques.
Furthermore, equivalence verification can be deployed to examine the influence of certain NN-based pre-processing steps (cf.~\cite{narodytska2018verifying}) or in cases where performing multiple verification tasks on a large NNs would be too expensive (cf.~\cite{Buning2020}).
In what follows, we refer to the current method of Paulsen et al.~\cite{paulsen2020reludiff} by \textsc{ReluDiff} and that of Kleine Büning et al.~\cite{Buning2020} by \textsc{MilpEquiv}.

One approach which has previously been shown to yield good results for adversarial and functional verification on NNs is Geometric Path Enumeration (GPE)~\cite{tran2019star,bak2020improved}.
However, this algorithm was initially devised as an approach operating on a single NN.
In this work we extend GPE to a setting with multiple NNs and implement its extension for the problem of equivalence verification.
We explore which (sometimes previously used) optimizations yield good results when applied to the equivalence problem.
While our work in this paper is specific to the problem of equivalence verification, the extended GPE algorithm can also be used for other verification tasks involving multiple NNs.
%Additionally, we show a theoretical result on coNP-completeness and perform a comparative evaluation of 3 equivalence verification tools (\textsc{ReluDiff}, \textsc{MilpEquiv} and our \textsc{NNEquiv}).
%Furthermore, we explore which optimizations yield good results in this modified setting.
%Additionally, we show a theoretical result on the coNP-completeness of the $\epsilon$-equivalence problem and perform a comparative analysis showing strengths and weaknesses of our approach.

\subsection{Contribution}
In this work, we focus on the problem of equivalence verification for (potentially) structurally differing NNs.
Our contributions are as follows:
\begin{itemize}
    \item[(C1)] We prove that the $\epsilon$-equivalence problem for NNs is coNP-com\-plete.
    %A proof that the problem of disproving $\epsilon$-equivalence is NP-complete.
    \item[(C2)] %We present a modified version of the GPE algorithm (Tran et al.~\cite{tran2019star}) and apply it to the equivalence verification problem.
    We extend the GPE algorithm (Tran et al.~\cite{tran2019star}) to a setting with multiple NNs and apply it to the equivalence verification problem.
    %An extension of the Geometric Path Enumeration (GPE) algorithm (previously presented by \cite{tran2019star}) to the problem of equivalence verification.
    \item[(C3)] We evaluate several optimizations for this setting which increase efficiency on practical problems.
    %A discussion of the necessary optimizations for the approach to be efficient in practice.
    \item[(C4)] We perform a comparative evaluation of our algorithm (on ACAS Xu and modified MNIST benchmarks) and show that it outperforms \textsc{MilpEquiv} in four of our five $\epsilon$-equivalence benchmarks and in counterexample finding.
\end{itemize}

\subsection{Overview}
The structure of our paper is as follows:
In \Cref{sec:prelim-related}, we present related work in the field of NN equivalence and introduce the basic notions of GPE relevant to our paper.
Before explaining the idea behind our extension of GPE to multiple NNs in \Cref{sec:extending}, we show that the $\epsilon$-equivalence problem is coNP-complete in \Cref{sec:np-complete}.
%Subsequently, in \Cref{sec:extending}, we explain how GPE can be extended to multiple NNs and can be instrumented for equivalence verification.
Starting from a naive algorithm, we then evaluate optimizations to enable efficient equivalence verification using GPE in \Cref{sec:optimizing}.
%Based on the naive algorithm, we then present the optimizations necessary for efficient equivalence verification using GPE in \Cref{sec:optimizing}. \todo[inline]{what is the naive algorithm?}
We further explore these 
%necessary 
optimizations, in particular the question of good refinement heuristics, in \Cref{sec:tree-explore}.
Finally, in \Cref{sec:eval}, we evaluate our algorithm and show advantages and disadvantages to the current state of the art represented by \textsc{MilpEquiv}
\cite{Buning2020}.

\section{Preliminaries \& Related Work}
\label{sec:prelim-related}
%In this section, we will briefly review works relevant to our research.
%In this section we will give a short introduction to various equivalence properties before presenting current state-of-the-art approaches in equivalence verification.
%Subsequently, we will give an introduction of Geometric Path Enumeration for single NNs that will be extended in later chapters.

%\todo[inline]{mkb: we should probably give short introduction for NNs and RELU }

\subsection{Feed-Forward NNs}
NNs consist of interconnected units called neurons.
% I believe this formulation might be easier for the GPE case since this corresponds to the structure of Star Sets and Zonotopes...
A neuron $j$ computes a non-linear function of its input values $x_1, \dots, x_n$ 
according to $y_j = \sigma(\sum_{i=1}^{n} w_{ij} x_i + b_j)$
where $\sigma$ is called the activation function and $w_{ij}$ are the weights.
$b_j$ is commonly referred to as bias of neuron $j$.
In this paper, we focus on the \emph{rectified linear unit} activation function,
$\relu(x) = \max(0, x)$, which is one of the most commonly used activation functions in modern NNs~\cite{Goodfellow-et-al-2016}. 
Outputs of neurons are connected as input to other neurons, resulting in a directed graph.
In this paper, we focus on feed-forward NNs, where the underlying graph is acyclic.
Neurons are organized in layers, where neurons in layer $l$ take inputs only from the directly preceding layer $l-1$.
The first layer---called input layer---is just a place holder for the inputs to be fed into the NN, the subsequent layers are called hidden layers, while the last layer---the output layer---holds the function value computed by the NN.
We refer to the input space dimension as $I\in\mathbb{N}$ and to the output dimension as $O\in\mathbb{N}$.

\subsection{Verifiying NN Equivalence}
There has recently been a line of work which proposes various compression techniques for NNs (for a full review see Cheng et al.~\cite{cheng2020survey}).
While such techniques have been shown to be useful in practice, most lack a formal proof of correctness and only  rely on empirical evidence.
The usage of such techniques thus raises the question of how to prove that two NNs $\mathcal{R}$ (reference) and $\mathcal{T}$ (test) and their corresponding mathematical functions $g_{\mathcal{R}}:\mathbb{R}^I\to \mathbb{R}^O$, $g_{\mathcal{T}}:\mathbb{R}^I\to \mathbb{R}^O$ are equivalent, i.e. that they produce the same results.
%In this context, we are particularly interested in $\relu$ networks which consist of linear layers which are interleaved by \textit{rectified linear units} ($\relu$s).

%we aim to answer the question whether the two networks are equivalent.
%It thus becomes necessary to precisely define what it means for two NNs to be equivalent in order to check such properties.
%To this end, 
Kleine Büning et al.~\cite{Buning2020} examined three possible definitions of equivalence
on a given subset of inputs $\mathcal{I}\subseteq \mathbb{R}^I$,
two of which we review here:%\todo{Add top k?}
\begin{definition}[$\epsilon$-Equivalence \cite{Buning2020} / Differential Equivalence~\cite{paulsen2020reludiff}]
\label{def:epsilon-equiv}
Two NNs $\mathcal{R}$ and $\mathcal{T}$ are $\epsilon$-equivalent with respect to a norm $\lVert \cdot \rVert$, if $\left\lVert g_{\mathcal{R}}\left(x\right)-g_{\mathcal{T}}\left(x\right)\right\rVert < \epsilon$ for all $x \in \mathcal{I}$
\end{definition}
%\todo{mkb: this property is checked by ReluDiff and called differential verification of NNs by them}
\begin{definition}[Top-1-Equivalence \cite{Buning2020}]
\label{def:one-hot-equiv}
Two NNs $\mathcal{R}$ and $\mathcal{T}$ are top-1-equivalent, if $\argmax_i r_i = \argmax_j t_j$ where $r=g_{\mathcal{R}}\left(x\right)$ and $t=g_{\mathcal{T}}\left(x\right)$ for all $x \in \mathcal{I}$
\end{definition}
%\todo{Alternative for indices in argmax?}
%\Cref{def:one-hot-equiv} can be further relaxed by considering more than one label:
%\begin{definition}[Top-$k$ Equivalence\cite{Buning2020}]
%\label{def:top-k-equiv}
%Two networks $\mathcal{R}$ and $\mathcal{T}$ are top-$k$ equivalent, if given $g_{\mathcal{R}}\left(x\right)=r$ and $g_{\mathcal{T}}\left(x\right)=t$:
%\[
%\argmax_i r_i = j \implies \pos(t_j, t) \leq k
%\]
%where $\pos(v_j, v)$ returns $i$, if $v_j$ is the $i$-th largest value of $v$
%\end{definition}
While Definition \ref{def:epsilon-equiv} is particularly suitable for regression tasks as it can show a very strong form of equivalence, the latter definition is more relaxed and especially useful for classification tasks.
%Paulsen et al.~\cite{paulsen2020reludiff} refer to the property in Definition \ref{def:epsilon-equiv} as \textit{differential equivalence}.

Paulsen et al.~\cite{paulsen2020neurodiff} and Kleine Büning et al.~\cite{Buning2020} proposed two fundamentally different approaches to NN equivalence verification. % with differing application areas.
While Paulsen et al.~\cite{paulsen2020neurodiff} proposed a technique called \textsc{ReluDiff}/\textsc{NeuroDiff} which uses (symbolic) interval propagation on NNs with similar weight configurations (e.g. produced through float truncation)
%structurally similar NNs (i.e. networks with the same architecture but differing weights)
to prove $\epsilon$-equivalence, the \textsc{MilpEquiv} technique \cite{Buning2020} is based on mixed integer linear programming (MILP) and encodes (potentially) structurally different NNs together with the desired equivalence property into an optimization problem.
\textsc{ReluDiff}/\textsc{NeuroDiff} was shown to be efficient for cases where a NN's weights had been truncated.
However, this approach does not work at all for structurally differing NNs and suffers in performance when weight differences are larger.
For such NNs \textsc{MilpEquiv} was shown to work well for a number of small instances.
%and all previously mentioned equivalence properties
(e.g. two 64 pixel input MNIST NNs with a total of 84 $\relu$ nodes).
In particular, \textsc{MilpEquiv} is able to provide a maximal radius around a datapoint for which the property still holds.

\subsection{Geometric Path Enumeration}
\label{subsec:related-gpe}
GPE is a methodology originally proposed by Tran et al.~\cite{tran2019star} for verifying safety properties in NNs.
Given a NN $\mathcal{N}$, a set of input instances $\mathcal{I}\subseteq\mathbb{R}^I$ and an unsafe output specification $\mathcal{U}\subseteq\mathbb{R}^O$ defined as a set of linear constraints,
safety verification is concerned with the question whether there exist any instances $i\in\mathcal{I}$ such that $g_{\mathcal{N}}\left(i\right) \in \mathcal{U}$.

% \begin{algorithm}
%     \caption{High-level path enumeration algorithm for NNs as described in \cite{bak2020improved}}
%     \begin{algorithmic}
%     \label{algorithm:gpe-basics:high-level}
%     \REQUIRE Input Set $\mathcal{I}$, Unsafe Set $\mathcal{U}$
%     \ENSURE Verification result (\texttt{safe} or \texttt{unsafe})
%     \STATE $s \leftarrow \left\langle \texttt{layer}:0, \texttt{neuron}: \texttt{None}, \Theta: \texttt{convert}\left(\mathcal{I}\right) \right\rangle$
%     \STATE $\texttt{W} \leftarrow \texttt{List}()$ \COMMENT{List of set datastructures to process}
%     \STATE $\texttt{W}.\texttt{put}\left(s\right)$
    
%     \WHILE{$\neg\mathcal{W}$.\texttt{empty}()}
%         \STATE $s \leftarrow \texttt{W}.\texttt{pop}()$
%         \STATE $\texttt{step}\left(s, \texttt{W}\right)$
%         \COMMENT{This call may modify \texttt{W} and $s$}
%         \IF{$\neg \texttt{finished}\left(s\right)$}
%             \STATE $\texttt{W.push}\left(s\right)$
%         \ELSIF{$\neg \texttt{verify}\left(s, \mathcal{U}\right)$}
%             \RETURN \texttt{unsafe}
%         \ENDIF
%     \ENDWHILE
%     \RETURN \texttt{safe}
%     \end{algorithmic}
% \end{algorithm}

Instead of pushing single data points through the NN and checking whether they satisfy the required safety property, GPE feeds an entire set into the NN and then evaluates whether any parts of the output sets lie inside $\mathcal{U}$.
The sets are represented through generalized star sets or zonotopes which we define below:
\begin{definition}[Generalized Star Set \cite{tran2019star}]
A generalized star set $\Theta$ is a tuple $\left\langle c, G, P\right\rangle$ where $c\in\mathbb{R}^n$ is the center, $G=\left(g_1\cdots g_m\right)\in\mathbb{R}^{n \times m}$ is the generator matrix, and $P\subseteq\mathbb{R}^m$ is a set defined through a conjunction of linear constraints (i.e. a polytope).
%and $P:\mathbb{R}^m \to \left\{\top,\bot\right\}$ is a predicate which is restricted to be a conjunction of linear constraints.
%\todo{Sometimes also refered to as half space polytope if we prefer that name?}
The set represented by $\Theta$ is then defined as:
\[
\left\llbracket \Theta \right\rrbracket = \left\{ x \in \mathbb{R}^n \;\middle\vert\; \exists \alpha \in P \,:\, x = c + G\alpha \right\}
\]
%We will sometimes refer to $G$ as generator and as matrix $G=\left(g_1g_2\cdots g_m\right)$.
%For simplicity we denote $c + G\alpha$ as $\Theta\left(\alpha\right)$.
%For simplicity we will sometimes refer to $c + G\alpha$ as $\Theta\left(\alpha\right)$.
\label{def:gpe:star-set}
\end{definition}
%\todo[inline]{Do we need generator and matrix representation?}
Alternatively, it is possible to use zonotopes which further restrict the type of predicate allowed:
\begin{definition}[Zonotopes \cite{bak2020improved}]
A zonotope $\Psi$ is a generalized star set $\langle c, G, P \rangle$ with the further restriction that $P$ may only be defined through interval constraints (i.e. $P$ only enforces a lower and upper bound for each dimension).
\end{definition}
%\todo{Zonotopes and Generalized Star Sets im Text immer klein}

We refer to zonotopes and generalized star sets as \emph{set data structures}.
Initially, the GPE algorithm converts the provided input space into either of these data structures and then propagates the sets through the NN.
The transformation deployed through the dense layer of a NN can be exactly represented by zonotopes and generalized star sets through application of the weight matrices to $G$ and $c$.
$\relu$ nodes require a different type of transformation since they are only piece-wise linear functions:
To this end, the data structures can either be split by introduction of an additional hyperplane to the linear constraint predicate (\textit{exact GPE}) or the $\relu$ function can be over-approximated~\cite{10.1007/978-3-030-76384-8_2,Singh2018} (\textit{approximate GPE}).
Note that optimization for generalized star sets is much more expensive than optimization for zonotopes, which can be computed with a closed form solution.

%\Cref{algorithm:gpe-basics:high-level} can be completed by providing suitable \texttt{step}, \texttt{finished} and \texttt{verify} functions for the desired data structure as previously done by \textcite{tran2019star}.

%While affine transformations can trivially be applied to both data structures, it is less clear how $\relu$ nodes should be handled by the datastructure.
%To this end, one can either split the set into two cases or one can use various over-approximation techniques\cite{Singh2018,10.1007/978-3-030-76384-8_2}.
%We call the approach which splits generalized star sets \textit{exact GPE} and other approaches \textit{approximate GPE} since Zonotopes are usually unable to provide an exact output.

\section{NN Equivalence and NP completeness}
\label{sec:np-complete}
\newcommand{\epsnetequiv}{\textsc{$\epsilon$-Net-Equiv}}
\newcommand{\netvery}{\textsc{Net-Verify}}
Katz et al.~\cite{katz2017reluplex} have previously shown that the satisfiability problem for linear input and output constraints of a single NN with $\relu$ nodes is NP-complete. We refer to this decision problem as \netvery.
In this section, we show that the $\epsilon$-equivalence problem for NNs
is coNP-complete. Since disproving $\epsilon$-equivalence is NP-complete, the task of proving $\epsilon$-equivalence is coNP-complete.

\begin{theorem}[\epsnetequiv{} is NP-complete]
Let $\mathcal{R},\mathcal{T}$ be two arbitrary ReLU NNs and let $\mathcal{I}$ be some common input space of the two NNs.
Determining whether $\exists x \in \mathcal{I}: \lVert g_\mathcal{R}\left(x\right)-g_\mathcal{T}\left(x\right)\rVert_p \geq \epsilon$ is NP-complete for any p-norm $\lVert \cdot \rVert_p$.
\end{theorem}
%Due to space constraints
The full proof can be found in \Cref{apx:npcomplete}.
%however we will briefly outline the idea of the proof at this point:
In essence, the proof consists of a reduction from \netvery{} to \epsnetequiv{}.
In order to reduce a \netvery{} instance consisting of a NN $\mathcal{N}$ and a linear constraint specification $\psi$, we encode it as follows:
The first NN $\mathcal{R}$ only consists of $\mathcal{N}$.
The second NN $\mathcal{T}$ consists of $\mathcal{N}$ and a suitable encoding of the linear constraints $\psi$.
We then show that we only can disprove $\epsilon$-equivalence iff $\mathcal{N}$ satisfies the given specification $\psi$.

\section{Extending GPE to multiple NNs}
\label{sec:extending}
%We extend the previously presented GPE approach to be applicable to multiple NNs.
The most trivial approach to extend GPE to multiple NNs would be to \textit{stitch} multiple NNs into a single composite NN and then execute regular GPE on this composite NN.
However, the composite NN's weight matrices would be considerably larger which would increase the computational load.
Furthermore, NNs with a different number of layers would have to be padded for this approach.
This would nullify any performance gains which could otherwise be achieved through the reduced NN size.

Instead, we propose to propagate star sets through both NNs sequentially.
By carefully selecting the constraint sets of the propagated sets, we can ensure that there remains a point-wise correspondence between the output data structures of GPE for the two (or more) NNs considered.
To make our approach clear, we introduce \emph{transfer functions} as a way of reasoning about exact propagation of set data structures.
\begin{definition}[Transfer Function]
Let $\mathcal{N}$ be a NN.
A transfer function $T_\mathcal{N}$ is a function which, given an input data structure $\Theta=\langle c,G,P \rangle$, produces a set of output data structures s.t.
%that represents all outputs of $\mathcal{N}$ for inputs in $\llbracket\Theta\rrbracket$, such that
%
\[
\forall \langle c',G',P' \rangle \in T_\mathcal{N}\left(\Theta\right) \;.\; \forall \alpha \in P' \;.\; g_\mathcal{N}\left(c+G\alpha\right) = c'+G'\alpha
\]
and that the union of all $P'$ within $T_\mathcal{N}\left(\Theta\right)$ equals $P$.
% \[
% \left\{g_\mathcal{N}\left(x\right) \mid x \in \llbracket \Theta \rrbracket \right\} \;\subseteq\; \bigcup_{O\in T_\mathcal{N}\left(\Theta\right)} \llbracket O \rrbracket
% \]
% We call $T_\mathcal{N}$ \textit{exact} if the two sides are equal. Otherwise $T_\mathcal{N}$ is an over-approximation.
\end{definition}

Using these transfer functions, we show that there is a correspondence between the output sets of two NNs in GPE:

\begin{theorem}[NN Output Correspondence]
    \label{thm:output-correspondence}
    Let $\mathcal{R},\mathcal{T}$ be two NNs with their corresponding transfer functions $T_\mathcal{R}$, $T_\mathcal{T}$ and let $\Theta=\langle c, G, P \rangle$ be some input data structure.
    For any $\Theta_\mathcal{R} = \langle c_\mathcal{R}, G_\mathcal{R}, P_\mathcal{R} \rangle \in T_\mathcal{R}\left(\Theta\right)$ and $\Theta_\mathcal{T} = \langle c_\mathcal{T}, G_\mathcal{T}, P_\mathcal{T}\rangle \in T_\mathcal{T}\left(\langle c, G, P_\mathcal{R}\rangle\right)$: \vspace{-2ex}
    \begin{multline*}
    \left\{ \left( g_\mathcal{R}\left(x\right), g_\mathcal{T}\left(x\right)\right) \mid x \in \llbracket \langle c, G, P_\mathcal{T} \rangle \rrbracket \right\}
    %=\\
    %\left\{ \left(c_\mathcal{R}+G_\mathcal{R}\alpha , c_\mathcal{T} + G_\mathcal{T}\alpha\right) \mid \exists \alpha \in \mathbb{R}^m P_\mathcal{T}\left(\alpha\right)\right\}
    =\\
    \left\{ \left(c_\mathcal{R}+G_\mathcal{R}\alpha , c_\mathcal{T} + G_\mathcal{T}\alpha\right) \mid \exists \alpha \in P_\mathcal{T}\right\}
    \enspace.
    %\llbracket\langle c^o_1, G^o_1, P^o_2 \rangle\rrbracket \times \llbracket\Theta^o_2\rrbracket
    \end{multline*}
    %The two sides are equal iff $T_\mathcal{R}$ and $T_\mathcal{T}$ are exact.
\end{theorem}
% \begin{proof}
% %Let $x\in\llbracket\langle c,G,P \rangle\rrbracket$ be an arbitrary input point.
% Let $\Theta=\llbracket\langle c,G,P \rangle\rrbracket$ be some input data structure.
% By definition, for every $\Theta_\mathcal{R}=\langle c_\mathcal{R}, G_\mathcal{R}, P_\mathcal{R} \rangle \in T_\mathcal{R}\left(\Theta\right)$ and for all $\alpha_\mathcal{R} \in P_\mathcal{R}$ we have that $g_\mathcal{R}\left(c+G\alpha_\mathcal{R}\right) = c_\mathcal{R}+G_\mathcal{R}\alpha_\mathcal{R}$.
% Moreover for every $\Theta_\mathcal{T}=\langle c_\mathcal{T}, G_\mathcal{T}, P_\mathcal{T}\rangle \in T_\mathcal{T}\left(\langle c,G,P_\mathcal{R}\rangle\right)$ and all $\alpha_\mathcal{T} \in P_\mathcal{T}$ we get that
% $g_\mathcal{T}\left(c+G\alpha_\mathcal{T}\right)=c_\mathcal{T}+G_\mathcal{T}\alpha_\mathcal{T}$.
% Since $P_\mathcal{T} \subseteq P_\mathcal{R}$ we further have
% $g_\mathcal{R}\left(c+G\alpha_\mathcal{T}\right)=c_\mathcal{R}+G_\mathcal{R}\alpha_\mathcal{T}$.
% The last two equalities correspond exactly to the equalities between the two components of the considered tuples. 
% Since this equality holds for all $\alpha \in P_\mathcal{T}$ we obtain the result stated in the theorem.
% \end{proof}
% This will be the easiest way to include over-approximation: Otherwise we will have to introduce projections of polytopes (dimensionalty changes) or will have to artificially split up the result of T which is both ugly and lengthy...
An over-approximation would produce additional, spurious points in the output of $T\left(\Theta\right)$ and may therefore produce spurious output tuples.
In this case the right side of \Cref{thm:output-correspondence} becomes a superset.
%, it will never remove points of the exact version.
This in turn gives rise to the modified GPE algorithm outlined in \Cref{algorithm:gpe-equiv:high-level}.
We begin by feeding our input data structure $\langle c, G, P\rangle$ into the first NN.
The propagation step function for the data structures (\texttt{step}) is the same as in the single NN GPE algorithm in \Cref{subsec:related-gpe}.
For every output star set $\langle c_\mathcal{R}, G_\mathcal{R}, P_\mathcal{R} \rangle$,
we restrict the input data structure according to the predicate of the output of the first NN, i.e. $\langle c, G, P_\mathcal{R} \rangle$.
Then we feed this data structure into the second NN to obtain $\langle c_\mathcal{T}, G_\mathcal{T}, P_\mathcal{T} \rangle$.
%we feed the input data structure with restricted predicate according to the output of the first NN, i.e. $\langle c, G, P_\mathcal{R} \rangle$, into the second NN and obtain $\langle c_\mathcal{T}, G_\mathcal{T}, P_\mathcal{T} \rangle$.
In the end, we can compare the two output tuples $\langle c_\mathcal{R}, G_\mathcal{R}\rangle$ and $\langle c_\mathcal{T}, G_\mathcal{T} \rangle$ constrained by the predicate $P_\mathcal{T}$.

Note that both considered output sets are therefore constrained by $P_\mathcal{T}$ (not $P_\mathcal{R}$).
This is the essential insight, that allows our approach to produce point-wise correspondences between the outputs of the two NNs.

\begin{algorithm}
    \caption{High-level path enumeration algorithm for equivalence checking. $^{LP}$ indicates the step uses LP solving.}
    \label{algorithm:gpe-equiv:high-level}
    \begin{algorithmic}
    \REQUIRE Input $\Theta = \langle c, G, P \rangle$, NNs $\langle\mathcal{R},\mathcal{T}\rangle$
    \ENSURE Verification result (\texttt{equiv} or \texttt{nonequiv})
    \STATE $s \leftarrow \langle \texttt{nn}:\mathcal{R}, \texttt{layer}:0, \texttt{neuron}: \texttt{None},$\\$\qquad\Theta: \Theta,\Theta_\mathcal{R}:\bot, \Theta_\mathcal{T}:\bot \rangle$
    \STATE $i \leftarrow \Theta$
    \STATE $\texttt{W} \leftarrow \texttt{List}()$ \COMMENT{Working list of set data structures}
    \STATE $\texttt{W}.\texttt{put}\left(s\right)$
    \WHILE{$\neg\texttt{W}$.\texttt{empty}()}
        \STATE $s \leftarrow \texttt{W}.\texttt{pop}()$
        \STATE $\texttt{step}_{s.\texttt{nn}}\left(s, \texttt{W}\right)^{LP}$ \COMMENT{Propagate $s.\Theta$ by one neuron}
        \IF{$s$ finished network $\mathcal{R}$}
            \STATE $s.\Theta_\mathcal{R} \leftarrow s.\Theta$ \COMMENT{Store output from $\mathcal{R}$ for comparison}
            \STATE $s.\texttt{nn}\leftarrow\mathcal{T}$
            \STATE $s.\Theta \leftarrow \langle i.c, i.G, s.\Theta.P \rangle$
            \STATE $s.\texttt{layer}, s.\texttt{neuron} \leftarrow 0, \texttt{None}$
        \ENDIF
        \IF{$s$ finished network $\mathcal{T}$}
            \STATE $s.\Theta_\mathcal{R}.P \leftarrow s.\Theta.P$ \COMMENT{Output of $\mathcal{R}$}
            \STATE $s.\Theta_\mathcal{T} \leftarrow s.\Theta$ \COMMENT{Output of $\mathcal{T}$}
            \IF{$\neg\texttt{is\_equiv}\left(s.\Theta_\mathcal{R},s.\Theta_\mathcal{T}\right)^{LP}$}
                \RETURN \texttt{not equivalent}
            \ENDIF
        \ELSE
            \STATE $\texttt{W.push}\left(s\right)$
        \ENDIF
    \ENDWHILE
    \RETURN \texttt{equivalent}
    \end{algorithmic}
\end{algorithm}

%Notice, that this scheme allows the consideration of multiple NNs, however, this work will be limited to the case of two NNs.

\subsection{Equivalence on Set Data Structures}
For our equivalence verification approach it is necessary to define an equivalence check \texttt{is\_equiv} which verifies whether two set data structures $\Theta_\mathcal{R},\Theta_\mathcal{T}$ satisfy $\epsilon$-equivalence or top-1 equivalence.
First, we present how $\epsilon$-equivalence with Chebyshev norm $\lVert \cdot \rVert_\infty$ can be proven for zonotopes.
Afterwards, we show how Star Sets can be used to prove $\epsilon$-equivalence and top-1 equivalence.

\paragraph{$\epsilon$-Equivalence}
In order to prove $\epsilon$-equivalence with the Chebyshev norm, we need to bound the maximum deviation between the two NN outputs by $\epsilon$.
That is, given the two output zonotopes $\Psi_\mathcal{R} = \langle c_\mathcal{R}, G_\mathcal{R}, P_\mathcal{T} \rangle$ and $\Psi_\mathcal{T} = \langle c_\mathcal{T}, G_\mathcal{T}, P_\mathcal{T} \rangle$ we want to find the maximal deviation:
\begin{align*}
    &\max_{\alpha \in P_\mathcal{T}} \lVert\left(c_\mathcal{R}+G_\mathcal{T} \alpha\right) - \left(c_\mathcal{T} + G_\mathcal{T} \alpha\right)\rVert_\infty\\
    = &\max_{i} \max_{\alpha \in P_\mathcal{T}} \left|\left(c_\mathcal{R}-c_\mathcal{T}\right)_i + \left(G_\mathcal{R}-G_\mathcal{T}\right)_i \alpha\right|.
\end{align*}
As can be seen by the reformulation above, we can find the maximal deviation over the output by solving optimization problems for each dimension of the \textit{differential zonotope}
\[
\partial\Psi = \langle \left(c_\mathcal{R}-c_\mathcal{T}\right), \left(G_\mathcal{R}-G_\mathcal{T}\right), P_\mathcal{T} \rangle).
\]
Recalling that zonotopes can be optimized with a closed form solution, this enables a quick check for the adherence of the desired $\epsilon$-equivalence property.
However, since zonotopes only approximate the output set, one may need to fall back to the use of Star Sets if equivalence cannot be established using zonotopes.
In this case, we can reuse the same formula from above to obtain a \textit{differential star set} $\partial\Theta$ which is then optimized using LP solving.

\paragraph{Top-1 Equivalence}
For top-1 equivalence there are two possible approaches which both rely on propapagated star sets.
We can reuse the \textsc{MilpEquiv} encoding and employ a MILP solver.
%For this approach, we modified the formulation used by \textsc{MilpEquiv} such that the problem becomes infeasible if the two NNs are equivalent.
%We can thus simply check whether the MILP instance provided to the solver is returned to be infeasible.
%If so, the top-1 property holds on the given output sets, otherwise a counterexample is returned.
Alternatively, we can use a simplex (LP) solver.
In the latter case we split up the output star set $\Theta_\mathcal{T}$:
%For every output dimension $j\in O$ we can generate a polytope $P_j$, which ensures that $j$ is always the maximum of $\langle c_\mathcal{R}, G_\mathcal{R}, P_\mathcal{T} \cap P_j\rangle$.

For every output dimension $1\leq j \leq O$ we generate a polytope $P_j$.
Additional constraints $r_j \geq r_i ~\forall i \neq j$ ensure that output $r_j$ is the maximum among the outputs of $\mathcal{R}$ in $\langle c_\mathcal{R}, G_\mathcal{R}, P_\mathcal{T} \cap P_j\rangle$.
%$P_j$ is defined through $O-1$ constraints where each constraint ensures that one particular output is less or equal to output $j$.
Note that the union of $P_1$ to $P_O$ covers all of $P_\mathcal{T}$.
We then examine the outputs of $\Theta_j=\langle c_\mathcal{T}, G_\mathcal{T}, P_\mathcal{T}\cap P_j\rangle$ for every $1 \leq j \leq O$.
Since $j$ is always the maximum of $\mathcal{R}$ for this part of the output space, we want to ensure that $j$ is also always the maximum of $\mathcal{T}$.
% For every of these $O$ set data structures, we optimize the difference between output $j$ and the output of any other dimension.
% For every output dimension $j \in O$ we produce a star set that is constrained in such a way that $j$ is the maximum of $\Theta_\mathcal{R}$.
% This is done through the introduction of an additional $O-1$ hyperplanes which ensure this property.
% % thus enabling us to establish top-1 equivalence.
% Then, for an $O$ dimensional output, we solve $O-1$ optimization problems per output dimension $j$.
% To this end, we generate $O-1$ constraints which ensure that dimension $j$ is the maximum in $\Theta_\mathcal{R}$ and add those constraints to our second output $\Theta_\mathcal{T}$.
Therefore, we compute the maximal difference between output dimension $j$ and the other dimensions in $\Theta_j$.
If all of these differences are below $0$, we can guarantee top-1 equivalence.
This procedure produces $\mathcal{O}\left(O\right)$ star sets and $\mathcal{O}\left(O^2\right)$ optimization operations in total.
%\todo[inline]{More explicit explanation of what is actually happening...}
%Top-k equivalence follows trivially by allowing a certain number of dimensions above $0$.
%\todo{top-k not introduced before}

\subsection{Challenges and Limitations of the approach}
While the techniques outlined above permit a straightforward extension of GPE to multiple NNs, and thus allow achieving equivalence verification, the approach comes with a number of pitfalls which should be avoided.
The most obvious is probably the possibility of exponential growth in the number of star sets.
As previously noted, the exact GPE approach based on star sets splits the star sets on ReLU nodes. 
%and can thus yield a high rate of exponential growth.
Tran et al.~\cite{tran2019star} rightly note that the observed growth usually drastically falls behind the worst case, however the increase in ReLU nodes through the processing of two NNs at once certainly leads to an increase in necessary splits.
This is particularly the case for ReLU nodes which cut off very similar hyperplanes (such as the two ReLU nodes in a NN at the same position with truncated weights).
This can not only double the work, but it may also lead to precision problems with LP solvers which tend to show problematic behavior when encountering a problem which has a very small feasible set\footnote{In one case the solver would return drastically differing maximum values for the same optimization problem depending on the previous requests or would suddenly deem the problem infeasible.}.
To avoid such numerical problems we thus use 64-bit floats by default and always ensure that feasibility is checked at least once by an exact (i.e. rational) LP solver before a branch is declared infeasible.
While this can mitigate most numerical problems,
the approach is weaker than \textsc{ReluDiff}/\textsc{NeuroDiff} for the specific use case of weight truncation for structurally similar NNs (e.g. truncation from 32-bit to 16-bit floats) -- this is best left to the approach presented by Paulsen et al.~\cite{paulsen2020neurodiff}.
Although these initial improvements help in making GPE for equivalence \textit{possible}, this approach is not yet \textit{scalable}.
Hence, we devote the next section to various optimizations.
%which we applied to improve the scalability of the approach.
% \begin{itemize}
%     \item Basic GPE algorithm for two networks
%     \item Equivalence Check for $\epsilon$ and top 1 equivalence
%     \item The problems you immediately observe
%     \begin{itemize}
%         \item Increased exponential growth of star sets due to (at worst) doubling of ReLU nodes
%         \item Floating Point problems which:
%         \begin{itemize}
%             \item Make it desirable to have an exact solver as fallback?
%             \item Make it desirable to not use too much LP min/max
%         \end{itemize}
%     \end{itemize}
% \end{itemize}

\section{Optimizing GPE for two NNs}
\label{sec:optimizing}
%While the approach presented above does allow equivalence verification of small NNs, this approach is not yet scalable.
The approach presented above is not yet scalable.
In particular, we identify two bottlenecks: The number of splits and the time taken for LP optimization.
%, this approach is not yet scalable as the necessary LP optimizations and the number of considered splits for the exact case becomes too big to handle in a reasonable time frame.
Therefore, we consider a number of optimizations, some of which have previously been used by Bak~\cite{10.1007/978-3-030-76384-8_2}.

\subsection{Zonotope Propagation}
As an initial optimization we reused the zonotope propagation technique presented by Bak et al.~\cite{bak2020improved}, which reduces the number of LP optimizations necessary through a zonotope based bounds computation.
We refer to this first version of the algorithm as \textsc{NNEquiv-E} (for exact).
As can be seen in \Cref{fig:version-runtimes} later on, this approach produces a total runtime of 54,390s on our 9 benchmark instances.

% The first, most obvious, optimization is the use of over-approximating zonotopes instead of exact star sets.
% To this end, we reuse the propagation algorithm previously proposed by \textcite{bak2020improved} which splits the zonotope on $\relu$ nodes overlapping $0$ and contracts the zonotope bounds based on the hyper plane that is currently being split.
% We initially explored a version of this algorithm which exclusively used zonotopes for bounds computation on $\relu$ splits and only checked feasibility after splitting through execution of the first simplex phase on the propagated star set, however it turned out that computing the bounds of split candidates exactly using the propagated star set is more efficient.
% This is, because copying and and further constraining LPs comes at a significant cost and the feasibility-only approach increased the number of star sets that had to be handled by the approach.
% We will refer to this version as \textsc{NNEquiv-E}.
% As can be seen in \Cref{fig:version-runtimes} later on, this approach produces a total runtime of 54390s on our 9 benchmark instances.

\subsection{Zonotope Over-Approximation}
%Even with these initial optimizations the algorithm is still left with a large number of sets which have to be evaluated.
%We thus have two options.
To further optimize the algorithm we can either reduce the time spent per zonotope or we can try to reduce the number of zonotopes which have to be considered.
In order to achieve the second objective, we can over-approximate certain $\relu$ splits through a methodology first presented by Singh et al.~\cite{Singh2018} and later reused by Bak~\cite{10.1007/978-3-030-76384-8_2}: 
The idea is to introduce an additional dimension to the zonotope and use it to over-approximate the $\relu$ node by a parallelogram.
%The idea is to introduce an additional dimension to the zonotope and use it to form a parallelogram around the area covered by the $\relu$ node.
Over-approximation errors accumulate across layers (Bak~\cite{10.1007/978-3-030-76384-8_2} refer to this as \textit{error snowball}).
To make the parallelogram as tight as possible and minimize the over-approximation error, we use the bounds computed through LP solving (instead of the looser zonotope bounds) if there were any exact splits beforehand.
%While this decreases the amount of zonotopes to consider, it still produces an over-approximation error.
%Since the over-approximation of all $\relu$ nodes can result in a situation where the desired properties are no longer provable due to approximation error, it becomes necessary to find a suitable refinement heuristic which decides what nodes are refined if an equivalence property cannot be established.
In an abstraction-refinement setting, we would start by propagating over-approximating zonotopes through both NNs and then check, whether the equivalence property can be established.
If the property does not hold, we refine one over-approximated $\relu$ node by splitting the zonotope and propagating the split zonotopes further through the NNs.

\begin{figure}
    \centering
    \includegraphics[width=\columnwidth]{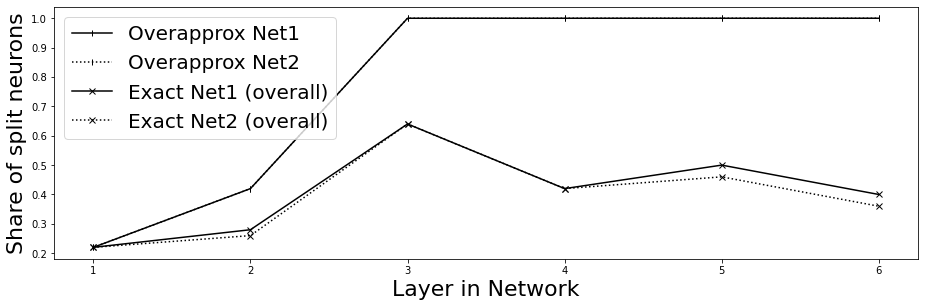}
    \caption{
    Comparison of exact star set propagation (Exact) and propagation of an over-approximating zonotope (Overapprox) for $\epsilon$-equivalence over ACAS\_1\_1: Net1 corresponds to $\mathcal{R}$ and Net2 corresponds to $\mathcal{T}$ (Overapprox Net2 is hidden behind Overapprox Net1)}
    \label{fig:gpe-splitting}
\end{figure}

%This happens, because the over-approximation increases poss
In \Cref{fig:gpe-splitting} we compare the share of nodes per layer whose bounds contain the value $0$ for an exact approach in comparison to the propagation of an over-approximating zonotope.
Any such node can be considered a split candidate which could be used for refinement.
Each refinement can then help in reducing the over-approximation error and in establishing the desired property.
%are ever split or over-approximation for exact propagation in comparison to over-approximating propagation for a proof of equivalence on an ACAS Xu NN:
As can clearly be seen in the plot, the over-approximation approach produces a lot more split candidates than the exact approach.
Not all of the splits candidates encountered for the over-approximation would actually have to be refined in the worst case.
This is, because many of the split candidates are only artifacts of previous over-approximations.
We refer to these split candidates as \textit{ghost splits}.
These \textit{ghost splits} cannot be easily distinguished from actual, necessary splits.
%The only split of which we know that it must exist, is the first split that is being over-approximated.
%Therefore, it is
The only guaranteed non-ghost split, is the first split candidate encountered, while all later split candidates might be artifacts of over-approximation.
%Note that this comparison aggregates the number of splits over all splitted sets for the exact case -- for a single propagated set the number of splits is usually even lower.
%This over-approximation can in turn result in even more over-approximation error which \textcite{10.1007/978-3-030-76384-8_2} referred to as \textit{error snowball}.

%We will discuss the search for a suitable refinement strategy in the next section, however there is one more efficiency improvement related to the dimensionality of the LP problems solved.
Thus, the simplest refinement strategy would be to refine only this node.
%the first over-approximated node in the NN.
We refer to this strategy as \textsc{NNEquiv-F} (First), and it reduces the runtime on our benchmark set to 2,489s (c.f. \Cref{fig:version-runtimes}).
However, this approach still leaves %considerable
room for improvement as we explain in \Cref{sec:tree-explore}.

\subsection{LP approximation}
With the introduction of over-approximation we encounter an additional problem:
Splitting hyperplanes are no longer dependent on the input variables \textit{only}, but also depend on the dimensions introduced through the over-approximation.
This raises the question how to handle the additional dimension in the propagated star set:
Since equally increasing the dimensionality of the LP problem leads to increased solver runtimes,
%increases the runtime of the solver
we instead opted to over-approximate the LP problem.
Classically, for an $m$ dimensional zonotope with initial input dimensionality $I$ we observe a hyperplane cut of the following form:
\begin{align}
\sum_{i=1}^I g_i \alpha_i + \sum_{i=I+1}^m g_i \alpha_i \leq c
\label{eq:exact-lp}
\end{align}
We can now over-approximate this inequality by computing $\mu = \min_{\alpha} \sum_{i=I+1}^m g_i \alpha_i$ through zonotope optimization and constraining the LP problem with the following inequality:
\begin{align}
    \sum_{i=1}^I g_i \alpha_i \leq c - \mu
    \label{eq:approx-lp}
\end{align}
Since any solution for \Cref{eq:exact-lp} implies a solution for \Cref{eq:approx-lp} the second inequality is an over-approximation and can be used to reduce the number of dimensions the LP solver has to handle despite the %use of 
over-approximation of the zonotope.% side.
Note that we need to take this over-approximation into account for minimization/maximization tasks.
Since the LP solver only optimizes the first $I$ dimensions, we need to add the optimization result of the over-approximating zonotope for the remaining dimensions.
%by adjusting the result of the LP solver with the minimal/maximal value of the other dimensions computed through the zonotope.
We refer to this version as \textsc{NNEquiv-A} (for approximate LP).
%As can be seen in 
\Cref{fig:version-runtimes} shows that this approach % further
reduces the runtime to 1,631s.

\section{The Branch Tree Exploration Problem}
\label{sec:tree-explore}
Given the introduced over-approximations over $\relu$ splits, it becomes necessary to define a strategy that decides which over-approximations are refined if it turns out that the property cannot be established with the current over-approximation.
The problem of refinement heuristics has previously been studied for single NNs by Bak~\cite{10.1007/978-3-030-76384-8_2} who experimentally showed that a classic refinement loop approach which over-approximates everything and step by step refines over-approximations starting at the beginning of the NN (i.e. \textsc{NNEquiv-F/A}) sometimes performs worse than exact analysis.
While we were able to reproduce this problem for some benchmark instances, we observed an improvement for others.
%the exact behavior for \textsc{NNEquiv-F/A} seems to be benchmark dependent.
% Instead of always refining the first encountered $\relu$ node we also explored various other refinement heuristics which tried to decrease the necessary time by making a better decision about which node to split.
% However, it turned out that all strategies suffered from at least one of the problems:
% \begin{itemize}
%     \item The computation of the decision is too costly and can thus not reduce solving times
%     \item The decision procedure cannot distinguish between actual split neurons and ghost splits which are only a result of previous over-approximations.
% \end{itemize}
It seems like a good approach to begin propagation with an exact strategy which splits on every encountered neuron, which, however, eventually transitions into over-approximation.

We proceed with a formal analysis on different strategies and their (dis)advantages.
For this we consider binary trees that are implicitly explored by a GPE algorithm:
For given NNs and input space $\mathcal{I}$, the implicit tree explored by GPE consists of vertices $V=N \cupplus L$ where $N$ are the inner nodes of the tree representing $\relu$ splits and $L$ are the leafs of the tree representing the output set data structures.
The execution of an exact GPE algorithm implicitly produces a set of paths of the form $p\in N^*\times L$ that are (for now) explored sequentially.
We denote this set of paths as $P$.
For the exact case, the number of explored paths is fixed to the number of leafs.
Since GPE produces a partitioning of the input space $\mathcal{I}$, we can associate a part of the input space to every leaf and to every inner node.
%$v \in L$ as $A\left(v\right)\subseteq\mathcal{I}$ and to every inner node $w\in N$ as $A\left(w\right)=A\left(r\left(w\right)\right)\cup A\left(l\left(w\right)\right)$ where $l,r$ are the left and right children of $w$.
For its execution GPE needs to \texttt{descend} into each leaf and execute a \texttt{check} function on each leaf to prove equivalence.
\texttt{descend} refers to the operations necessary to process a star set up to its next $\relu$ split.
\texttt{check} refers to the operations necessary to prove equivalence on an output star set.
Since the \texttt{descend} function is executed once for each of the $2\left|P\right|-2$ edges of the tree and the \texttt{check} function is executed once for each of the $\left|P\right|$ leaves, the execution time of \textsc{NNEquiv-E} is bounded by
$\mathcal{O}\left( \left|P\right|*\left(t_{\texttt{check}} + 2t_{\texttt{descend}}\right) \right)$.

Omitting the option of reordering the inner nodes and thus producing a smaller tree, we must either  reduce $t_{\texttt{check}}$ and $t_{\texttt{descend}}$ or $\left|P\right|$ to reduce solving times.
In many cases, the considered property $\mathcal{E}$ cannot only be proven on part of the input space associated to the leaf, but there also exists some inner node $n \in N$ with an associated part of the input space which is already sufficiently partitioned to show $\mathcal{E}$ using over-approximation.
For a given equivalence property $\mathcal{E}$ we can define a function $\min_{\mathcal{E}}:N^*\times L \to \mathbb{N}$ which returns the number of necessary steps in the given path $p$ for the property $\mathcal{E}$ to be verifiable on the input space part associated to element $\min_{\mathcal{E}}\left(p\right)$ of path $p$.
The exploration of the tree induced by the set of paths
\[
P^*=\left\{ p' \in N^* \cup N^*\times L \;|\; \exists p \in P: p'=p_{1 : \min_{\mathcal{E}}\left(p\right)}\right\}
\]
would then be sufficient to prove equivalence ($p_{1:\min_{\mathcal{E}}\left(p\right)}$ denotes the prefix path of $p$ from step $1$ to step $\min_{\mathcal{E}}\left(p\right)$).

\begin{figure}
    \centering
    \includegraphics[width=1\columnwidth]{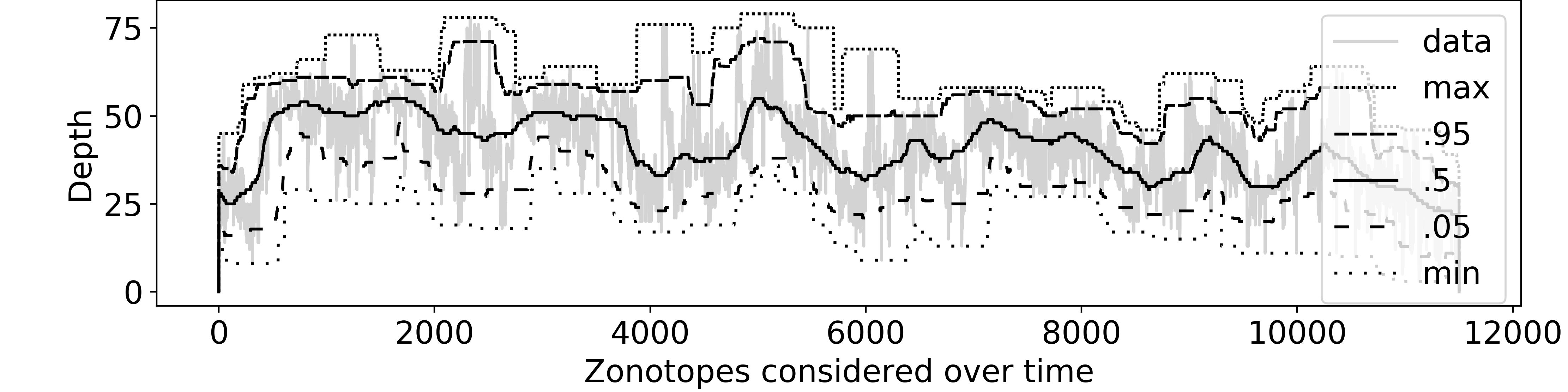}
    \caption{ACAS\_1\_1-retrain: Descend depth at point of successful equivalence proof for $\epsilon=0.05$ (Running percentile window width: 479)}
    \label{fig:success-depth}
\end{figure}

\textsc{NNEquiv-F/A} manages to obtain this minimal number of paths -- however at the cost of a much higher time spent on each path.
In particular, \texttt{check} is not only executed for each leaf, but also for each inner node.
Ignoring the over-approximation costs, this produces the following lower bound for the cost of \textsc{NNEquiv-F/A}:
\[
\Omega\left( \left|P^*\right|*\left(2t_{\texttt{check}} + 2t_{\texttt{descend}}\right)\right)
\]
Even when assuming the omitted over-approximation steps to be free, \textsc{NNEquiv-F/A} becomes less effective than \textsc{NNEquiv-E} if asymptotically $\left|P\right| < \left|P^*\right|\frac{2\left(t_{\texttt{check}}+t_{descend}\right)}{t_{\texttt{check}}+2t_{\texttt{descend}}}$, i.e. if the reduction of paths in $P^*$ is insignificant in comparison to the check time for the additional paths.
While there are cases where \textsc{NNEquiv-F/A} is effective, this is not guaranteed to be the case -- especially for larger NNs with higher values for $\min_\mathcal{E}$ (which increases $\left|P^*\right|$) and expensive check functions.

However, this formal framework allows us to define the (virtual) optimal run which takes the minimal amount of work for a given tree:
An algorithm which has an oracle for $\min_{\mathcal{E}}$ and always over-approximates at the right node.
This approach has a runtime of
$\mathcal{O}\left( \left|P^*\right|*\left(t_{\texttt{check}} + 2t_{\texttt{descend}}\right)
\right)$.

Since $\left|P^*\right| \leq \left|P\right|$ and the omitted over-approximation time tends to be smaller than the \texttt{descend} time, this approach can provide the optimum achievable through %suitable
heuristics for $\min_{\mathcal{E}}$.
In fact, we simulated such virtual runs using a pre-computed oracle by computing $\min_\mathcal{E}$ using \textsc{NNEquiv-A} and descending only the minimum necessary number of steps for each path.
In our evaluation we refer to this approach as \textsc{NNEquiv-O}.
As expected, % from our analysis,
\textsc{NNEquiv-O} produced the best results of all variants considered in our work running only 635s on our benchmark set.
This is not a practical algorithm, but provides a lower bound for the time taken using $\min_\mathcal{E}$ heuristics.

%It is thus important to find a good trade-off between the number of paths to consider and the number of operations necessary per path.
It is thus important to find a good heuristic which estimates $\min_{\mathcal{E}}$.
These heuristics are much more difficult to analyze theoretically because they are particularly dependent on the distribution of the encountered paths.
%and would thus only allow a probabilistic analysis.
%for which we would then have to assume a probability distribution on the paths.
%To this end we explored two heuristics for $\min_{\mathcal{E}}$ which show that heuristics can both increase and decrease efficency.
Therefore, we only explore two heuristics experimentally which show that heuristics have a significant impact on the runtime.
%which show that heuristics can increase and decrease efficiency.

\Cref{fig:success-depth} plots the depth at which GPE was successful in proving equivalence for a path in an ACAS Xu NN (i.e. the values of $\min_\mathcal{E}$).
Besides the data in grey, we plotted a number of running percentiles over the depth values.

A strategy which we have found to be inefficient is the use of a running maximum over the number of refinements needed by previous paths. %which previous runs needed for establishing the property.
This strategy is referred to as \textsc{NNEquiv-M} (for maximum) and drastically increases runtime to 19,191s,
presumably by over-estimating 
%This is because \textsc{NNEquiv-M} tends to over-estimate 
the number of refinements, thus increasing the number of paths considered.

Since \Cref{fig:success-depth} suggest that there are \textit{phases} in which the NN needs deeper or less deep refinement depths, we considered a heuristic which predicts a refinement depth equal to the depth of the previous path minus 1.
%While 
This accounts for the possible phases of the depth and also ensures that the algorithm is \textit{optimistic} in the sense that it always tries to reduce the number of refinement steps.% necessary which in turns, if successful, reduces the number of considered paths.
This can then reduce the number of considered paths.
We refer to this heuristic as \textsc{NNEquiv-L} which reduces runtime on the benchmark set by another 5\% to 1,553s.
While the methodology of over-approximation using Zonotopes is the same for \textsc{NNEquiv-A} and \textsc{NNEquiv-O/L/M} the approaches differ in the strategy deciding where the over-approximation is refined.

%{\color{red} \textbf{OLD}}
%\input{sections2/branch_tree_exploration.tex}

\section{Experimental Evaluation}
\label{sec:eval}
To evaluate our approach, we implemented the GPE based equivalence verification technique using parts of a pre-existing (single NN) GPE implementation by Bak et al.~\cite{bak2020improved} in Python.
We will refer to our implementation as \textsc{NNEquiv}\footnote{Our implementation is available on GitHub: \url{https://github.com/samysweb/nnequiv}}.
Our evaluation aims at answering the following questions:
\begin{itemize}
    \item[(E1)] Do the proposed optimizations make the algorithm more efficient?
    \item[(E2)] How does \textsc{NNEquiv} compare to previous work such as \textsc{MilpEquiv}~\cite{Buning2020} and \textsc{ReluDiff}~\cite{paulsen2020reludiff}?\footnote{Unfortunately, there is no artifact for \textsc{NeuroDiff}~\cite{paulsen2020neurodiff} which we could have evaluated.}
    \item[(E3)] How does the tightness of the $\epsilon$-equivalence constraint influence solving behavior? 
    %\item[(E4.1)] How does the use of the MILP top1 encoding influence the results?
    %\item[(E4)] Can \textsc{NNEquiv} help in the search of counter-examples for equivalence?
\end{itemize}

\begin{figure}
    \centering
    \includegraphics[width=\columnwidth]{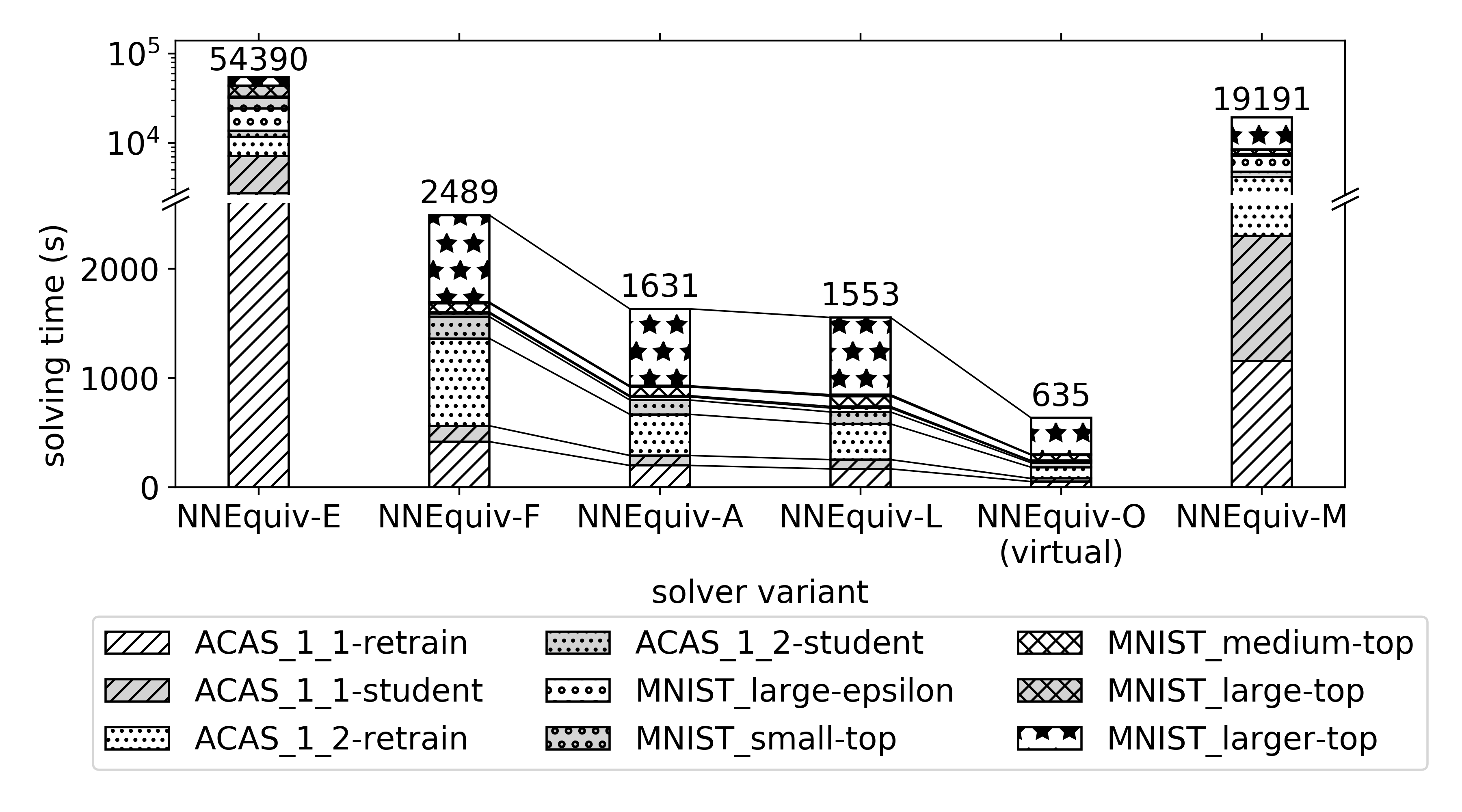}
    \caption{Time in seconds taken for our equivalent benchmarks per version. Note that the upper half of the y-axis has a logarithmic scale for improved visibility of the results.}
    \label{fig:version-runtimes}
\end{figure}
\subsection{Experimental Setup}
%To answer these questions it became necessary to find a suitable set of benchmarks to test the approaches on.
The benchmark landscape for the task of equivalence verification is still very limited.
Paulsen et al.~\cite{paulsen2020reludiff} proposed a number of benchmark NNs consisting of pairs of NNs differing only in the bit-width of the weights (32 bit vs. 16 bit).
As discussed before, we see this as a restricted use case and are more interested in generic NNs with varying structures and weights.
%with potentially drastic variations in structures and weight configurations.
This is %also the reason
why we omit a comparison on these NNs where the approach by Paulsen et al.~\cite{paulsen2020neurodiff}  is clearly faster and more precise.
Structurally differing NNs have been previously proposed by Kleine Büning et al.~\cite{Buning2020} who examined 3 NNs of differing layer depths for digit classification on an MNIST~\cite{lecun1998gradient} data set with reduced resolution~\cite{Dua:2019} (8x8 pixels).
%\todo{not MNIST}
% dataset taken from https://archive.ics.uci.edu/ml/datasets/Optical+Recognition+of+Handwritten+Digits

In order to %properly
evaluate and compare the approaches we thus proceeded as follows:
First, we decided to look at two types of NNs: Image classification on the 8x8 pixel MNIST data set and NNs used in control systems in the context of an Airborne Collision Avoidance System (ACAS Xu~\cite{julian2016policy}).
Then, based on the original ACAS Xu NNs 1\_1 and 1\_2, we contructed a total of 4 mirror NNs through retraining (ACAS\_1\_1-retrain, ACAS\_1\_2-retrain) and student teacher training~\cite{hinton2015distilling} for smaller NNs (ACAS\_1\_1-student, ACAS\_1\_2-student).
In additon to the smallest and largest MNIST 8x8 NNs considered in previous work (MNIST\_small-top, MNIST\_medium-top), we constructed two %even
larger MNIST models using student teacher training (MNIST\_large-top, MNIST\_larger-top).
Moreover, we constructed a second version of MNIST\_large-top for $\epsilon$-equivalence verification (MNIST\_large-epsilon).
All NNs were trained using variants of student teacher training and were trained in such a way that they were likely to be top-1 or $\epsilon$-equivalent in some parts of the input space.
More details on the properties of the 9 considered benchmark NNs are available online.\footnote{An overview table of all benchmarks is available at \url{https://github.com/samysweb/nnequiv-experiments/blob/main/benchmarks.md}}
%\todo{Maybe add link to GitHub here and put table of benchmarks there?}
%Finally, we had to decide on which share of the equivalent input space to evaluate the verification on.
The input space considered for verification is a sensitive choice, as it can have significant and varying impact on the performance of different verification techniques.
%Changes in the input can thus make an approach look better or worse.
For the case of GPE, the algorithm's performance tends to degrade with increasing input space size due to the growth in necessary splits.
Therefore, for each individual benchmark, we decided to look at an input size which was hard to handle for \textsc{NNEquiv-E}.
%the \textsc{NNEquiv-E} version was not at all able to handle within the given time frame or which required processing a significant number of star sets.
This has two reasons.
First, it allows us to evaluate the impact of the optimizations presented above in their ability to decrease runtimes.
Secondly, it permits to compare the performance of \textsc{NNEquiv} to the performance of \textsc{MilpEquiv} on instances which are difficult \textit{for our} approach.
The entire experimental setup can be found online.\footnote{On GitHub: \url{https://github.com/samysweb/nnequiv-experiments}}
%\todo{Drin lassen oder zu defensiv?}
%While one might still be concerned this choice of benchmarks could be biased, we draw attention to the fact that we do not outperform \textsc{MilpEquiv} in all instances allowing a more detailed comparison of the strengths and weaknesses of the two approaches and underlining our effort to achieve a fair comparison.

We used a machine with 4 AMD EPYC 7281 16-Core processors (i.e. 64 cores in total) and a total of 64GBs of RAM.
All experiments were run with a single thread, a memory limit of 512MB\footnote{The memory limit was irrelevant in practice, as no experiment hit this limit.}, and a timeout of 3 hours.
The experiments were run in parallel, up to 24 processes at once.
All times given in the subsequent sections are the median of 3 runs.

\subsection{Comparison of \textsc{NNEquiv} versions}
\label{subsec:experiments-versions}

%In a first step, we evaluated the impact of the previously outlined optimizations on the runtime of our algorithm.
\Cref{fig:version-runtimes} shows that the proposed optimizations help in reducing the runtime of the algorithm (note that the upper half of the y-axis has a logarithmic scale for improved visibility of the results).
%Note, that the bar partitioning in the Figure needs to be interpreted with care, as the y axis had to be scaled logarithmically.
%In particular, this figure permits to evaluate the impact of the proposed refinement heuristics.
On the one hand, we can observe, that heuristics for $\min_\mathcal{E}$ can, in principle, improve and worsen the result of the approach (as seen with \textsc{NNEquiv-L} and \textsc{NNEquiv-M}).
On the other hand, we see that there is still significant room for improvement through the development of better refinement heuristics -- this optimization would be supplementary to further optimizations which could be developed.
%\todo{Other visualisations?}

\subsection{Comparison to previous work}
\begin{table}[]
    \centering
    \caption{Runtime comparison (in seconds) for % our most efficient version 
    \textsc{NNEquiv-L} and \textsc{MilpEquiv}}
    \begin{tabular}{l|c|c|c}
    Benchmark & Property & \textsc{NNEquiv-L} &  \textsc{MilpEquiv} \\
    \hline\hline
    ACAS\_1\_1-retrain &$\epsilon=0.05$&   \textbf{167.45} &       TO \\
    ACAS\_1\_1-student &$\epsilon=0.05$&    \textbf{84.85} &       TO \\
    ACAS\_1\_2-retrain &$\epsilon=0.05$&   \textbf{326.59} &       TO \\
    ACAS\_1\_2-student &$\epsilon=0.05$&   \textbf{109.46} &         320.07 \\
 MNIST\_large-epsilon &$\epsilon=15$&    35.90 &          \textbf{19.97} \\
     MNIST\_small-top &top-1&    14.39 &           \textbf{3.51} \\
    MNIST\_medium-top &top-1&    94.51 &           \textbf{3.85} \\
     MNIST\_large-top &top-1&    \textbf{13.02} &          25.85 \\
    MNIST\_larger-top &top-1&   706.56 &         \textbf{386.04} \\
    \end{tabular}
    \label{tab:nnequiv-milpequiv}
\end{table}
The comparison to \textsc{MilpEquiv} is shown in \Cref{tab:nnequiv-milpequiv}.
\textsc{NNEquiv} outperforms \textsc{MilpEquiv} on the ACAS instances, where \textsc{MilpEquiv} even runs into a timeout for three of the four verification tasks.
In particular, this seems to be the case for larger NNs with low-dimensional inputs.
The superior performance of \textsc{MilpEquiv} for the case of MNIST\_large-epsilon seems to be caused by the LP solver in \textsc{NNEquiv} which is a magnitude slower for solving optimizations tasks for MNIST in comparison to ACAS Xu.
As this cannot be explained by the number of constraints, we suspect this is a problem related to the larger input dimensionality for the MNIST case (64 inputs in comparison to 5 inputs for ACAS Xu).
The ACAS-retrain NNs have the same structure as the original ACAS NN, allowing us to compare \textsc{NNEquiv} to \textsc{ReluDiff}.
While \textsc{ReluDiff} was able to quickly verify equivalence for the truncated NNs, where the mean absolute weight difference was $\approx 9.37*10^{-5}$, it was significantly slower than our approach on the retrain instances, with a mean weight difference of $\approx 0.48$, for $\epsilon \leq 5$ and even timed out for smaller values of $\epsilon$.
This suggests that the applicability of \textsc{ReluDiff} is not only restricted to structurally similar NNs, but that its performance also heavily depends on small weight differences.
%The results indicate that there is a class of $\epsilon$-equivalent NNs which can only be proven by \textsc{NNEquiv} or can be proven faster.
%\textsc{NNEquiv} can prove equivalence for some NNs where \textsc{MilpEquiv} is slower or times out.
%In particular, this seems to be the case for larger NNs with low-dimensional inputs.
%We also compared our approach to \textsc{ReluDiff} on the instances where it is applicable.
%\textsc{ReluDiff} timed out on the two ACAS-retrain instances.
%This is possibly due to the larger differences in the retrained NNs' weights in comparison to truncated NNs.
%Moreover, \textsc{ReluDiff} timed out on the two retrain instances which it can process due to their structural similarity.
%This makes \textsc{NNEquiv} the only approach which, to the best of our knowledge, is able to solve these ACAS instances within the timeout.
%solve the two retrain instances which \textsc{ReluDiff} can process due to their structural similarity.
%The faster speed of \textsc{MilpEquiv} for the case of MNIST\_large-epsilon seems to be caused by the LP solver in \textsc{NNEquiv} which is a magnitude slower for solving optimizations tasks for MNIST in comparison to ACAS Xu.
%As this cannot be explained by the number of constraints, we suspect this is a problem related to the larger input dimensionality for the MNIST case (64 inputs in comparison to 5 inputs for ACAS Xu).
Regarding question (E2) we note that our approach is applicable to a broader class of NNs than \textsc{ReluDiff} and solved instances where both other approaches timed out.
%outperforms \textsc{MilpEquiv} for networks with small input dimensions.
%is competitive to previous work and furthermore solves some instances which could not be solved before.
%While we are not faster than \textsc{MilpEquiv} on all NNs, in particular not for the case of top-1 equivalence, we are also not much worse:
%For all considered benchmarks we are able to solve the problem in a reasonable amount of time.

%- Für Epsilon Äquivalenz ist unser Ansatz in den meisten hier betrachteten Fällen besser, insb. für ACAS Xu Netzwerke.
% - Nachschauen: Bottleneck von MNIST\_large-epsilon
% Time per LP solving task is significantly slower in this case (one order of magnitude in comparison to ACAS\_1\_-retrain). As this cannot be explained by the number of constraints, we suspect this is a problem related to the larger input dimensionality for the MNIST case which the LP solver has to handle.
%- Slower than Kleine Büning, but still able to verify in reasonable time
%=> Probleme lösbar, die davor nicht lösbar waren und ansonsten auch in akzeptabler Zeit lösbar

\subsection{Influence of $\epsilon$-equivalence tightness}
\begin{figure}
    \centering
    \includegraphics[width=0.9\columnwidth]{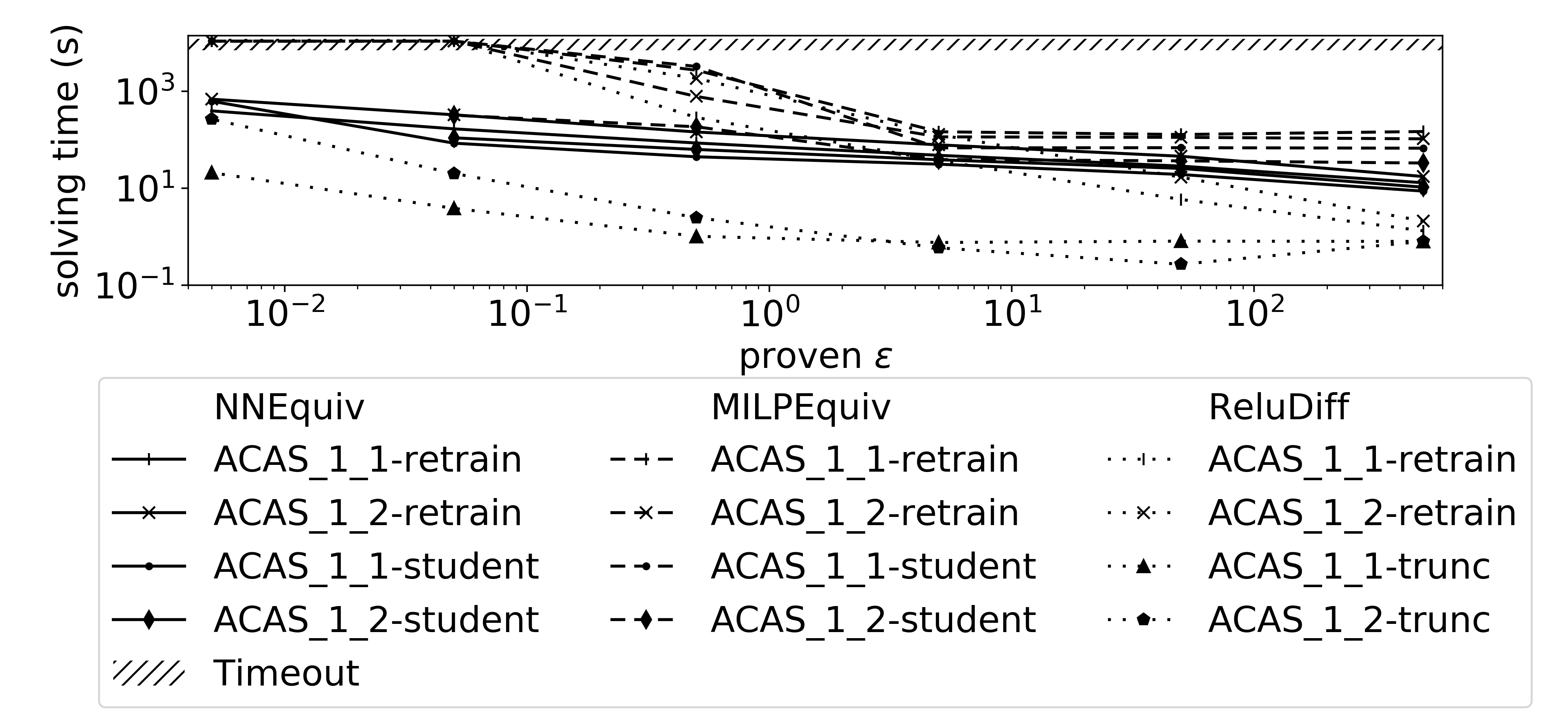}
    \caption{Solving time in seconds for $\epsilon$-equivalence with varying $\epsilon$: Solving times increase with tighter properties, however \textsc{NNEquiv-L} outperforms \textsc{MilpEquiv}. \textsc{ReluDiff} is outperformed by \textsc{NNEquiv-L} for retrained NNs.}
    \label{fig:epsilon-tightness}
\end{figure}
Concerning question (E3) we evaluated the performance of the approaches as we vary the tightness of $\epsilon$-equivalence for $\epsilon\in\left[0.005,500\right]$.
%We varied $\epsilon$ between 0.005 and 500 on 4 benchmarks and observed the changes in running time for the two approaches.
Note that we did not prove equivalence for $\epsilon=0.005$ for ACAS\_1\_2-student as we found this NN not to be 0.005-equivalent to the original NN.
Intuitively, a proof for a tighter $\epsilon$ bound will require more work as the approach either needs to refine more over-approximations (in the case of \textsc{NNEquiv-L}) or do further branch-and-bound operations (in the case of \textsc{MILPEquiv}).
We can observe this behavior in \Cref{fig:epsilon-tightness} which plots the runtime of \textsc{MILPEquiv} and \textsc{NNEquiv-L} as we tighten the $\epsilon$ bound.
Taking into account the log scales on both axes, we can observe that \textsc{NNEquiv-L} is at least one magnitude faster in proving equivalence for ACAS Xu NNs for $\epsilon\leq0.05$.
In particular, \textsc{MILPEquiv} produces time-outs for 3 of the 4 considered NNs once $\epsilon\leq0.05$.
We therefore suspect that our approach is better at handling very tight $\epsilon$ constraints in large NNs with low dimensional input.
This could potentially be due to the fact that GPE can use additional NN information (layer structure etc.) for its refinement decisions which is not readily available in the branch-and-bound algorithm in the backend of \textsc{MILPEquiv}.
For comparison, we plotted the performance of \textsc{ReluDiff} on NNs for truncated weights and retrained NNs:
As can be seen in \Cref{fig:epsilon-tightness} the approach by Paulsen et al.~\cite{paulsen2020reludiff} behaves similarly with respect to $\epsilon$ tightness.
Additionally, we see that the approach is less efficient for retrained NNs where the equivalence for $\epsilon\leq0.05$ cannot be established.
%\todo[inline]{highlight that retrained NNs have same structure, but different weights?}

% \todo{Rerun tightness experiment with \textsc{LAST-1}}
% 1. Je kleiner das zu beweisende Epsilon, und je genauer somit die Eigenschaft die bewiesen werden soll, desto mehr Zeit wird benötigt, da mehr Überapproximationen aufgelöst werden müssen
% 2. Bei \cite{Buning2020} ist dies  genauso, allerdings steigt für den Fall von ACAS Xu der Rechenaufwand deutlich schneller (Log Skala!)

\subsection{Finding Counterexamples}
Our technique can also be used to find counterexamples, showing that two NNs are not equivalent at a certain point.
This information can be interesting to further train NNs after a failed equivalence proof.
To this end, we compared the capabilities of \textsc{NNEquiv-L} in counterexample finding with the capabilities of \textsc{MilpEquiv}.
%While equivalent NNs are hard to come by, it is obviously very easy to find non-equivalent NNs.
%However, not all such cases must be hard to disprove.
To account for possibly easy instances, we looked at a large number of non-equivalent input spaces for each of our benchmark NNs which we know to be equivalent on other parts of the input space.
%Using brute force fuzzing, we extracted 100 equivalence counter-examples per benchmark NN.
To generate counterexamples, we randomly sampled input points and selected the points with differing NN outputs.
%We propagated these samples through both networks and checked the NN outputs on the benchmark's equivalence property.
Using this technique, we produced 100 distinct non-equivalent input points for each of our 9 benchmarks.
These points were used as center for $L_\infty$-balls which represented our input spaces.

We then evaluated \textsc{NNEquiv-L} and \textsc{MilpEquiv} on the same input space radii as in \Cref{subsec:experiments-versions} with the objective of finding counterexamples.
%While the input spaces may have overlapped partially, they were never the same thus yielding a different input structure every time which the approaches had to handle.
Since counterexample extraction is much faster than equivalence proofs, we set a timeout of 2 minutes. %per sample.
We found that \textsc{NNEquiv-L} was significantly faster and extracted a counterexample for 890 of the 900 considered benchmarks, while even a version without expensive $\relu$ bounds computation\footnote{While this optimization step improves performance for equivalence proofs, it may degrade performance for counterexample finding.} executed in the initialization (named \textsc{MilpEquiv-B}) was only able to find 315 counterexamples.
Also, the time per %found
counterexample was significantly lower for \textsc{NNEquiv-L}, making this approach an interesting technique for retraining NNs via counterexamples.
Looking at the behavior of \textsc{MilpEquiv}'s solver backend, it seems that the reason for our superior performance lies in the time needed by \textsc{MilpEquiv} to find an initial feasible solution.
\textsc{MilpEquiv} first needs to resolve the integer based $\relu$ node encodings, which are automatically resolved by \textsc{NNEquiv-L} through the propagation of sets.
%We suspect, that the superior performance by \textsc{Last-1} is caused by
%Concerning (E4) we remark that \textsc{Last-1} significantly outperf
\textsc{NNEquiv-L} has the potential to extract polytopes of non-equivalent input space subsets which could allow for even more efficient sampling.
% \begin{itemize}
%     \item Evaluation difficult, not so easy to find \textit{hard} counter examples
%     \item We therefore decided to evaluate this on a large number of input spaces per network.
%     \item Each input space had equal size to the ones used for equivalence verification
%     \item Centers of input spaces were found experimentally through brute force using uniformly random inputs
%     \item Input spaces may be overlapping, but are never the same
%     \item Total of 900 benchmarks (100 input spaces per network)
%     \item Counter-example finding is faster than proving equivalence. Therefore we set a timeout of 2 minutes per benchmark
% \end{itemize}
\begin{table}
    \centering
    \caption{Comparison of counterexample finding capabilities of \textsc{NNEquiv-L}, \textsc{MilpEquiv}  and \textsc{MilpEquiv-B} (no $\relu$ node bounds)}
    \begin{tabularx}{\columnwidth}{X|c|c|c}
         & \textsc{NNEquiv-L} & \textsc{MilpEquiv} & \textsc{MilpEquiv-B} \\\hline\hline
        \#Solved & \textbf{890} & 305 & 315\\
        Time~(incl.TO) & \textbf{3,597s} & 75,989s & 72,515s\\
        Time/Solved~(excl.TO) & \textbf{2.69s} & 14.91s & 7.21s\\
    \end{tabularx}
    \label{tab:my_label}
\end{table}

%Benchmarks:
% \begin{itemize}
%     \item top 1: Jeweils kleiner Bereich und mittel großer Bereich
%     \begin{itemize}
%         \item[A1] MNIST8x8 from \cite{Buning2020}
%         \item[A2] Largern MNIST 8x8
%         \item[B] MNIST8x8 PCA50
%         \item[C] MNIST784
%     \end{itemize}
%     \item Epsilon
%     \begin{itemize}
%         \item[D] ACAS Xu from \cite{paulsen2020neurodiff}
%         \item[E] ACAS Xu Student Teacher Training with random samples?
%         \item[F] MNIST Student Teacher?\\
%         Just had a look and the results do not look very promising (epsilon 20-30). It seems like Gurobi has an easier time if the required epsilon is larger
%     \end{itemize}
% \end{itemize}

\section{Conclusion and Future Work}
\label{sec:conclusion}
We proposed an approach extending Geometric Path Enumeration~\cite{tran2019star} to multiple NNs.
Employing this method, we presented an equivalence verification algorithm which was %subsequently
optimized by four techniques: Zonotope propagation, zonotope over-approximation, LP approximation, and refinement heuristics.
Our evaluation shows that the optimizations increase the approach's efficiency and that it can verify equivalence of NNs which were not verifiable by \textsc{MilpEquiv}~\cite{Buning2020} and \textsc{ReluDiff}~\cite{paulsen2020reludiff}.
Our approach significantly outperforms the state of the art in counterexample finding by solving 890 instances in comparison to 315 instances solved by \textsc{MilpEquiv}.
In addition, we proved the coNP-completeness of the $\epsilon$-equivalence problem, and presented a formal way of reasoning about refinement heuristics in the context of GPE.

%While this paper marks the beginning of the use of GPE for multiple NN verification, there are still a number of open questions.
%, some of which we will mention below.

In terms of efficiency, one could further explore possible refinement heuristics and consider parallelized (possibly GPU based) implementations.
Moreover, while GPE can increase the confidence in NNs, the role of numerical stability for the verification approach has to be further investigated.
%In terms of efficiency, there is a need to explore further refinement heuristics pushing the runtime of the algorithm closer to its optimum.
%Furthermore, a (possibly GPU based) parallelization could further improve the runtimes of the algorithm for equivalence proofs.
%
Furthermore, an integration of MILP constraints into GPE propagation could be explored resulting in an algorithm inbetween \textsc{NNEquiv} and \textsc{MilpEquiv}.
%More fundamentally, we have now explored two ends of a scale:
%On the one hand, the GPE approach, which decomposes the verification problem into a large number of small LP problems.
%On the other hand, the \textsc{MilpEquiv} approach which encodes the entire verification problem into one large MILP problem.
%It might thus be interesting to look at integrations of the two approaches which use a MILP instance instead of an LP problem as a propagated set.
%
Additionally, we see a need for a larger body of equivalence benchmarks which allows the conclusive evaluation of equivalence verification algorithms.

\bibliographystyle{IEEEtran}
\bibliography{IEEEabrv, main}

\begin{thebibliography}{10}
\providecommand{\url}[1]{#1}
\csname url@samestyle\endcsname
\providecommand{\newblock}{\relax}
\providecommand{\bibinfo}[2]{#2}
\providecommand{\BIBentrySTDinterwordspacing}{\spaceskip=0pt\relax}
\providecommand{\BIBentryALTinterwordstretchfactor}{4}
\providecommand{\BIBentryALTinterwordspacing}{\spaceskip=\fontdimen2\font plus
\BIBentryALTinterwordstretchfactor\fontdimen3\font minus
  \fontdimen4\font\relax}
\providecommand{\BIBforeignlanguage}[2]{{%
\expandafter\ifx\csname l@#1\endcsname\relax
\typeout{** WARNING: IEEEtran.bst: No hyphenation pattern has been}%
\typeout{** loaded for the language `#1'. Using the pattern for}%
\typeout{** the default language instead.}%
\else
\language=\csname l@#1\endcsname
\fi
#2}}
\providecommand{\BIBdecl}{\relax}
\BIBdecl

\bibitem{julian2016policy}
K.~D. Julian, J.~Lopez, J.~S. Brush, M.~P. Owen, and M.~J. Kochenderfer,
  ``{Policy compression for aircraft collision avoidance systems},'' in
  \emph{2016 IEEE/AIAA 35th Digital Avionics Systems Conference (DASC)}.\hskip
  1em plus 0.5em minus 0.4em\relax IEEE, 2016, pp. 1--10.

\bibitem{bojarski2016end}
M.~Bojarski, D.~Del~Testa, D.~Dworakowski, B.~Firner, B.~Flepp, P.~Goyal, L.~D.
  Jackel, M.~Monfort, U.~Muller, J.~Zhang, and {others}, ``{End to end learning
  for self-driving cars},'' \emph{arXiv preprint arXiv:1604.07316}, 2016.

\bibitem{katz2017reluplex}
G.~Katz, C.~Barrett, D.~L. Dill, K.~Julian, and M.~J. Kochenderfer,
  ``{Reluplex: An efficient SMT solver for verifying deep neural networks},''
  in \emph{International Conference on Computer Aided Verification}.\hskip 1em
  plus 0.5em minus 0.4em\relax Springer, 2017, pp. 97--117.

\bibitem{Singh2018}
G.~Singh, T.~Gehr, M.~Mirman, M.~P{\"{u}}schel, and M.~Vechev, ``{Fast and
  effective robustness certification},'' \emph{Advances in Neural Information
  Processing Systems}, vol. 2018-Decem, no. Nips, pp. 10\,802--10\,813, 2018.

\bibitem{wang2018efficient}
S.~Wang, K.~Pei, J.~Whitehouse, J.~Yang, and S.~Jana, ``{Efficient formal
  safety analysis of neural networks},'' in \emph{Advances in Neural
  Information Processing Systems}, 2018, pp. 6367--6377.

\bibitem{tran2019star}
H.-D. Tran, D.~M. Lopez, P.~Musau, X.~Yang, L.~V. Nguyen, W.~Xiang, and T.~T.
  Johnson, ``{Star-based reachability analysis of deep neural networks},'' in
  \emph{International Symposium on Formal Methods}.\hskip 1em plus 0.5em minus
  0.4em\relax Springer, 2019, pp. 670--686.

\bibitem{Buning2020}
M.~Kleine~B{\"{u}}ning, P.~Kern, and C.~Sinz, ``{Verifying Equivalence
  Properties of Neural Networks with ReLU Activation Functions},'' in
  \emph{Principles and Practice of Constraint Programming - 26th International
  Conference, CP 2020, Louvain-la-Neuve, Belgium, Proceedings}, 2020.

\bibitem{paulsen2020neurodiff}
B.~Paulsen, J.~Wang, J.~Wang, and C.~Wang, ``{NEURODIFF:} scalable differential
  verification of neural networks using fine-grained approximation,'' in
  \emph{35th {IEEE/ACM} International Conference on Automated Software
  Engineering, {ASE} 2020, Melbourne, Australia}.\hskip 1em plus 0.5em minus
  0.4em\relax {IEEE}, 2020, pp. 784--796.

\bibitem{narodytska2018verifying}
N.~Narodytska, S.~P. Kasiviswanathan, L.~Ryzhyk, M.~Sagiv, and T.~Walsh,
  ``{Verifying Properties of Binarized Deep Neural Networks},'' in \emph{AAAI},
  2018.

\bibitem{cheng2020survey}
Y.~Cheng, D.~Wang, P.~Zhou, and T.~Zhang, ``{A Survey of Model Compression and
  Acceleration for Deep Neural Networks},'' 2020.

\bibitem{paulsen2020reludiff}
B.~Paulsen, J.~Wang, and C.~Wang, ``Reludiff: differential verification of deep
  neural networks,'' in \emph{{ICSE} '20: 42nd International Conference on
  Software Engineering, Seoul, South Korea}, G.~Rothermel and D.~Bae,
  Eds.\hskip 1em plus 0.5em minus 0.4em\relax {ACM}, 2020, pp. 714--726.

\bibitem{bak2020improved}
S.~Bak, H.-D. Tran, K.~Hobbs, and T.~T. Johnson, ``{Improved Geometric Path
  Enumeration for Verifying ReLU Neural Networks},'' in \emph{International
  Conference on Computer Aided Verification}.\hskip 1em plus 0.5em minus
  0.4em\relax Springer, 2020, pp. 66--96.

\bibitem{Goodfellow-et-al-2016}
I.~Goodfellow, Y.~Bengio, and A.~Courville, \emph{Deep Learning}.\hskip 1em
  plus 0.5em minus 0.4em\relax MIT Press, 2016,
  \url{http://www.deeplearningbook.org}.

\bibitem{10.1007/978-3-030-76384-8_2}
S.~Bak, ``nnenum: Verification of relu neural networks with optimized
  abstraction refinement,'' in \emph{{NASA} Formal Methods - 13th International
  Symposium, {NFM} 2021, Proceedings}, A.~Dutle, M.~M. Moscato, L.~Titolo,
  C.~A. Mu{\~{n}}oz, and I.~Perez, Eds., vol. 12673.\hskip 1em plus 0.5em minus
  0.4em\relax Springer, 2021, pp. 19--36.

\bibitem{lecun1998gradient}
Y.~LeCun, L.~Bottou, Y.~Bengio, and P.~Haffner, ``{Gradient-based learning
  applied to document recognition},'' \emph{Proceedings of the IEEE}, vol.~86,
  no.~11, pp. 2278--2324, 1998.

\bibitem{Dua:2019}
\BIBentryALTinterwordspacing
D.~Dua and C.~Graff, ``{UCI} machine learning repository,'' 2017. [Online].
  Available:
  \url{https://archive.ics.uci.edu/ml/datasets/Optical+Recognition+of+Handwritten+Digits}
\BIBentrySTDinterwordspacing

\bibitem{hinton2015distilling}
G.~Hinton, O.~Vinyals, and J.~Dean, ``{Distilling the knowledge in a neural
  network},'' \emph{arXiv preprint arXiv:1503.02531}, 2015.

\end{thebibliography}

\appendix

\subsection{NN Equivalence and NP Completeness}
\label{apx:npcomplete}

\begin{theorem-nonum}[\epsnetequiv{} is NP-complete]
Let $\mathcal{R},\mathcal{T}$ be two arbitrary ReLU NNs and let $\mathcal{I}$ be some common input space of the two NNs.
Determining whether $\exists x \in \mathcal{I}: \lVert g_\mathcal{R}\left(x\right)-g_\mathcal{T}\left(x\right)\rVert_p \geq \epsilon$ is NP-complete for any p-norm $\lVert \cdot \rVert_p$.
\end{theorem-nonum}
\begin{proof}
Since all $p$-norms are equivalent up to some multiplicative factor, we show this theorem for $p=\infty$. $\epsilon$ and $d^*$ can be suitably modified for any other $p$.

We begin by showing that the problem is in NP.
%The problem is in NP.
Assuming a witness $x$ %returned by some algorithm 
for a given instance of \epsnetequiv{}, we can easily check whether the witness is violating the $\epsilon$ equivalence property by computing $\lVert g_\mathcal{R}\left(x\right)-g_\mathcal{T}\left(x\right)\rVert_\infty$.

%We need to establish that \epsnetequiv{} is NP-complete.
Next, we demonstrate a reduction from \netvery{} to \epsnetequiv{}.
An instance of \netvery{} is given by a conjunction of linear constraints on the input ($\psi_1\left(x\right)$) and on the output ($\psi_2\left(y\right)$) as well as a NN $\mathcal{N}$.
$\psi\left(x,y\right)=\psi_1\left(x\right)\land\psi_2\left(y\right)$ is said to be satisfiable if there is an $x$ such that $\mathcal{N}$ returns $y$ for input $x$ and $\psi\left(x,y\right)$ holds.
Note that we can represent $\psi_1$ and $\psi_2$ as a matrices of linear constraints of the form:
\begin{align*}
    C_1 x \leq b_1 & & C_2 y \leq b_2
\end{align*}
Reusing $\mathcal{N}$ and denoting $g_\mathcal{N}\left(x\right)$ as the output of $\mathcal{N}$ for input $x$ we can then construct a NN with the following outputs:
\begin{align}
    \max\left(0,C_1 x - b_1+\epsilon\right) \label{apx:netout:inputc}\\
    \max\left(0,C_2 g_\mathcal{N}\left(x\right) - b_2+\epsilon\right)\label{apx:netout:outputc}\\
    g_\mathcal{N}\left(x\right)\label{apx:netout:netout}
\end{align}
\Cref{apx:netout:inputc} has as many dimensions as $\psi_1$ has constraints and \Cref{apx:netout:outputc} has as many dimensions as $\psi_2$ has constraints.
Note that outputs of both (\ref{apx:netout:inputc}) and (\ref{apx:netout:outputc}) are only larger $0$ if an assignment is closer than $\epsilon$ to a constraint or violates it.
%constraint is violated or an assignment is closer than $\epsilon$ to the bound imposed by the constraint.

We now make use of an additional $\relu$ gadget which computes the non-negative maximum of two values:
\[
\relumax\left(a,b\right) = \max\left(0, a + \max\left(0,b-a\right)\right)
\]
We can then construct a pyramid of $\relumax$ gadgets on top of (\ref{apx:netout:inputc}) and (\ref{apx:netout:outputc}) which outputs the maximum deviation $d_{max}$.
Note that this pyramid of maxima is polynomially bounded in the problem size.
$d_{max}>\epsilon$ iff there exists a constraint on in- or output which is violated.
This follows from the observation that $d_{max}>\epsilon$ iff the maximum value of (\ref{apx:netout:inputc}) and (\ref{apx:netout:outputc}) is larger than $\epsilon$.
(\ref{apx:netout:inputc}) and (\ref{apx:netout:outputc}) contain a value larger than $\epsilon$ iff there is some constraint $\left(c,b\right)$ and some input $i\in\left\{x,y\right\}$ such that:
\[
c i - b + \epsilon > \epsilon \iff c i > b
\]
Conversely, $d_{max}\leq\epsilon$ iff all constraints are satisfied.
We now define $d^* = \max\left(0, 2\epsilon-d_{max}\right)$ and our NN's final output is:
\begin{align}
g_\mathcal{N}\left(x\right)+d^* \label{apx:netout:final}
\end{align}
By checking $\epsilon$-equivalence on $\mathcal{N}$ and the output defined in \Cref{apx:netout:final} we can solve our original \netvery{} instance:
\begin{align*}
&\lVert\left(g_\mathcal{N}(x)+d^*\right)-g_\mathcal{N}(x)\rVert_\infty\geq\epsilon\\
\iff & \lVert d^*\rVert_\infty\geq\epsilon\\
\iff & 2\epsilon-d_{max}\geq\epsilon\\
\iff & \epsilon \geq d_{max}\\
\iff & \text{all constraints are satisfied}
\end{align*}
% If instead of looking at \lVert \cdot \rVert_1, we would keep the same d*, however we would modify $\epsilon' = n\epsilon$.
%This would yield:
%n*|d^*| >= n \epsilon
%2n\epsilon - n*d_max >= n\epsilon
%2\epsilon-n*d_max > \epsilon
%...
I.e. any input which violates $\epsilon$ equivalence, is a solution to the $\netvery$ instance.
It follows that \epsnetequiv{} is NP-complete.
\end{proof}

\end{document}